\documentclass[letterpaper]{article}
\pdfoutput=1  

\usepackage[T1]{fontenc}
\usepackage[utf8]{inputenc}
\usepackage{textcomp}            %

\usepackage{geometry}
\usepackage{setspace}

\usepackage{amsmath}
\usepackage{amssymb}
\usepackage{amsfonts}
\usepackage{mathrsfs}

\usepackage{multirow}
\usepackage{booktabs}
\usepackage{xcolor}
\usepackage{graphicx}
\usepackage{float}

\newlength{\figwidth}
\usepackage{authblk}

\usepackage[style=chem-acs,articletitle=true,chaptertitle=true,backend=biber]{biblatex}
\AtBeginBibliography{\sloppy}

\usepackage[hidelinks]{hyperref}
\hypersetup{bookmarksdepth=2}
\usepackage[all]{hypcap}

\title{iPINN for Broadband CARS Phase Retrieval: A Framework for
  Function Approximation and Inverse Modeling Problems in Nonlinear
  Spectroscopy}

\author[1,2]{Ravi Teja Vulchi}
\author[2]{Carl Messerschmidt}
\author[1,2]{Mohammadsadegh Vafaeinezhad}
\author[1]{Rajendhar Junjuri}
\author[2]{Tobias Meyer-Zedler}
\author[1,2]{Juergen Popp}
\author[1,2]{Thomas Bocklitz*}

\affil[1]{Institute of Physical Chemistry (IPC) and Abbe Center of
  Photonics (ACP), Friedrich Schiller University Jena, Member of the
  Leibniz Centre for Photonics in Infection Research (LPI),
  Helmholtzweg 4, 07743 Jena, Germany}
\affil[2]{Leibniz Institute of Photonic Technology, Member of Leibniz
  Health Technologies, Member of the Leibniz Centre for Photonics in
  Infection Research (LPI), Albert-Einstein-Strasse 9, 07745 Jena,
  Germany}

\date{*Email: thomas.bocklitz@uni-jena.de}

\begin{document}

\maketitle

\begin{abstract}
  Broadband coherent anti-Stokes Raman scattering (BCARS) records chemically specific vibrational spectra, but the measured signal mixes the resonant, Raman-like response of the sample with a non-resonant background that changes from one measurement to the next. Recovering the resonant response, known as phase retrieval, is an ill-posed inverse problem. We present iPINN, an inverse physics-informed neural network that predicts the positions, amplitudes, and widths of Lorentzian vibrational bands directly from a BCARS spectrum. A transformer network estimates these band parameters, and a differentiable analytical model of the resonant susceptibility rebuilds the Raman-like spectrum from them as a sum of physically defined bands. We train the network on simulated spectra in which the bands of each reference compound stay fixed while the shape and strength of the background and the noise vary. Training also encourages similar band predictions for the same compound across these variations. On a public benchmark of toluene and dimethyl sulfoxide spectra, the network reached a mean absolute error about one third of that of the best of five deep-learning baselines. Its error for dimethyl sulfoxide stayed nearly constant across three noise levels. We also used 28 spectra of seven solvents, recorded at four focal depths with different background strengths. For five of the seven solvents, the mean absolute error changed little with depth, while it varied more for the other two. These results indicate that this approach keeps phase retrieval stable for most of the tested solvents when the background changes.
\end{abstract}

\section*{Keywords}
Semi-supervised learning, Physics AI, Inverse modeling, Artifact
correction, BCARS spectroscopy

\section{Problem Statement and Motivation}\label{sec1}

Inverse problems that recover a sparse set of latent parameters from indirect measurements are ubiquitous in the physical sciences and are inherently ill-posed, which makes priors and regularization essential for stable inference~\cite{kaipio2005,vulchi2025}. The same difficulty arises throughout spectroscopy and related photonic measurement systems in artifact removal tasks, where the recorded signal is distorted by background interference or instrumental noise that obscures the underlying physical response~\cite{houhou2022,vulchi2025}.

Coherent anti-Stokes Raman scattering (CARS), and in particular its broadband form (BCARS), is a representative example of this challenge. Although BCARS enables chemically specific measurements, the recorded spectrum is distorted by interference between the resonant contribution and the non-resonant background (NRB), producing asymmetric line shapes that are not directly interpretable. The chemically meaningful Raman-like information is encoded in the complex third-order nonlinear susceptibility, $\chi^{(3)}$, particularly in the imaginary part (referred to here as the resonant spectrum) of its resonant component, which follows a Lorentzian spectral shape. However, this phase information is not measured directly and must instead be inferred from intensity-only observations~\cite{junjuri2025}.

A useful phase-retrieval method must therefore remain stable across substantial measurement variability. NRB shape, NRB strength relative to the resonant signal, instrumental broadening, and signal-to-noise ratio (SNR) all change between focal positions, samples, and acquisition setups, whereas the resonant spectrum is governed by a small, low-dimensional band set. The chemistry of interest is invariant, but the artifacts that affect it are not. The method must therefore separate this invariant molecular content from acquisition-dependent distortions rather than fit any single representative measurement condition.

\subsection{The Challenge of Inverse Modeling in CARS}\label{subsec1}

Because phase and susceptibility components are not directly observable, classical BCARS workflows rely either on experimental NRB-suppression strategies or on numerical phase-retrieval methods such as Kramers--Kronig (KK) and maximum entropy methods~\cite{liu2009,okuno2010}. Both methods typically require reference NRB spectra, or equivalent normalization standards, together with non-trivial parameter choices, and their performance degrades on spectra with finite spectral support, non-ideal spectral features, or substantial measurement noise~\cite{junjuri2025,liu2009,rinia2007}. KK retrieval in particular is theoretically defined over an infinite frequency range, so the truncation imposed by an experimental window introduces phase artifacts that often require additional preprocessing~\cite{liu2009,camp2016,camp2020}. Wavelet-based methods model a slowly varying background or error signal and remove it from the measured spectrum~\cite{kan2016,harkonen2023}. In impedance spectroscopy, the Gaussian process Hilbert transform evaluates the KK relations probabilistically and returns credible intervals for the reconstructed spectrum~\cite{ciucci2020}.

Deep learning (DL) alternatives, including convolutional neural networks (CNNs), recurrent neural networks, and physics-informed approaches, have been proposed to bypass these limitations~\cite{valensise2020,houhou2020,junjuri2022,junjuri2022oc,saghi2022,wang2022vector,junjuri2023,muddiman2023,harkonen2024,saghi2024,vernuccio2024,vemuri2025}. However, a recent review and a comparative study of CARS phase retrieval indicate that current DL pipelines inherit two persistent weaknesses~\cite{junjuri2025,junjuri2023}. First, they remain sensitive to the choice of synthetic training distribution and exhibit larger errors near spectral edges and for weak features. Second, validation is most often performed on a narrow set of acquisition conditions, so reported accuracy does not directly indicate cross-instrument generalizability. Public real-experimental benchmarks typically contain only a handful of spectra acquired under fixed acquisition settings~\cite{junjuri2023,muddiman2023}. Physics-informed networks have been applied to this problem in two recent forms~\cite{muddiman2023,vemuri2025}. One of them, RamPINN, enforces KK consistency through a differentiable Hilbert-transform loss~\cite{vemuri2025}. Like the other deep-learning methods above, both physics-informed networks estimate a gridded spectrum rather than a set of physical parameters, so peak positions and intensities are obtained from it by detection, and linewidths require a separate fit. These considerations motivate inverse models that predict physically meaningful parameters directly and remain stable across acquisition conditions.

\subsection{The Inverse Physics-Informed Neural Network (iPINN) Framework}\label{subsec1_2}

Our goal is to recover a compact, physically meaningful set of peaks from BCARS measurements rather than estimate an unconstrained spectrum on a fixed grid. We describe this compact description as the band set: a finite set $\theta = \{A_k, \omega_k, \Gamma_k\}_{k=1}^{K}$ of amplitudes, wavenumber centers, and linewidths that fully specifies the resonant susceptibility under a sum of Lorentzian models. This differs from the original physics-informed neural network setting, in which neural networks approximate continuous solution fields and, in inverse variants, a small number of coefficients of governing equations~\cite{raissi2019}. In BCARS, the unknown is instead naturally discrete and parametric: a small number of resonances are embedded in a measurement that the NRB dominates.

\begin{figure}[!htbp]
\centering
\includegraphics[width=\figwidth]{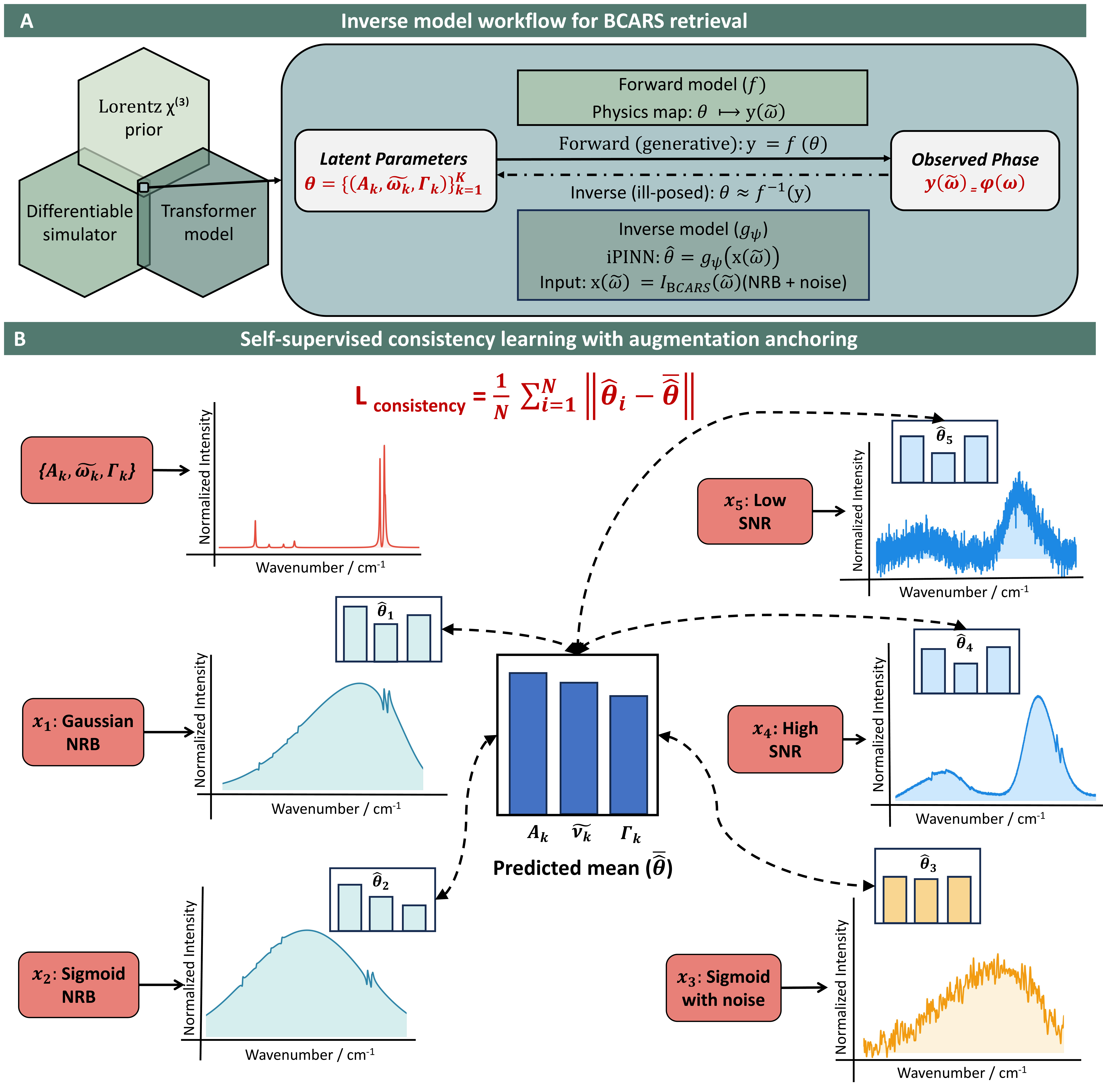}
\caption{Inverse modeling workflow for BCARS phase retrieval. (A)~A band set $\theta=\{A_k,\omega_k,\Gamma_k\}_{k=1}^{K}$ of amplitudes ($A_k$), wavenumber centers ($\omega_k$), and linewidths ($\Gamma_k$) defines the parametric prior $\chi^{(3)}_{\mathrm{prior}}$. The differentiable forward simulator implements the physics model $f$, mapping $\theta$ to the observable phase spectrum $y(\omega)=f(\theta)$. The inverse problem $\theta\approx f^{-1}(y)$ is ill-posed, and we approximate it with a transformer-based inverse model $g_{\psi}$ that predicts $\theta=g_{\psi}(x(\omega))$ from measured BCARS spectra $x(\omega)=I_{\text{BCARS}}(\omega)$, which include NRB and noise. (B)~Starting from a single band set $\theta$, multiple augmented BCARS variations are generated by varying NRB shape, NRB strength, and noise level. Each variation is passed through the inverse model to produce a prediction $\theta_i$, and a consistency loss penalizes disagreement among the $\theta_i$, encouraging the network to recover invariant physical parameters despite stochastic background and noise variation.}\label{fig:intro}
\end{figure}

We therefore adopt an inverse-first formulation: instead of mapping spectrum to spectrum, the network maps a measured BCARS spectrum directly to $\theta$, and a differentiable physics layer reconstructs the resonant spectrum from $\theta$ through the analytical Lorentzian model (Fig.~\ref{fig:intro}A). This places the Lorentzian shape constraint inside the architecture rather than in the loss alone, so features outside the predicted band set are suppressed by construction. The inverse model is implemented with transformer layers because self-attention integrates information across the full BCARS wavenumber axis~\cite{vaswani2017}. This is important for BCARS because the spectra contain three very different regimes in practice: weak but information-rich fingerprint bands below 1800~cm$^{-1}$, the largely feature-sparse silent region from about 1800 to 2800~cm$^{-1}$, and the typically stronger C--H stretch region between 2800 and 3100~cm$^{-1}$. Global context allows the model to stabilize local resonance estimates across regions and addresses generalization issues reported for CNN-based and shallow recurrent baselines, which can struggle with long-range dependencies and edge behavior on narrow synthetic training sets.

To keep the inverse problem well-grounded during training, supervision is introduced at the level of $\theta$ whenever reference information is available. Peak-center priors are anchored to tabulated Raman shift values from established reference standards, including ASTM E1840 and the NIST Chemistry WebBook for common solvents~\cite{astm_e1840,nist_webbook,hickstein2018,carrabba2006,nist_methanol,nist_ethanol}. The cyclohexane peak centers are taken from Carrabba's table of the values of Vickers and Mann~\cite{carrabba2006,vickers1999}. This gives the model physically grounded peak targets rather than synthetic labels with no chemical interpretation. Robustness is then improved through two additional constraints. The first is a physics-consistency objective implemented through the differentiable forward simulator, which regularizes $\theta$ so that the predicted band set reproduces the simulated resonant spectrum through the forward map. The second is a self-supervised consistency objective, which encourages the same $\theta$ to be recovered from multiple corrupted views of the same molecular sample (Fig.~\ref{fig:intro}B). The corrupted views are drawn from three NRB functional families (Gaussian, sigmoidal, and polynomial) combined with broad ranges of NRB strength and SNR. This work develops the framework and tests it on solvents.

\section{Datasets and Methods}\label{sec2}

\subsection{Procedural Data Generation Pipeline}\label{subsec2_1_1}

\begin{figure}[!htbp]
\centering
\includegraphics[width=\figwidth]{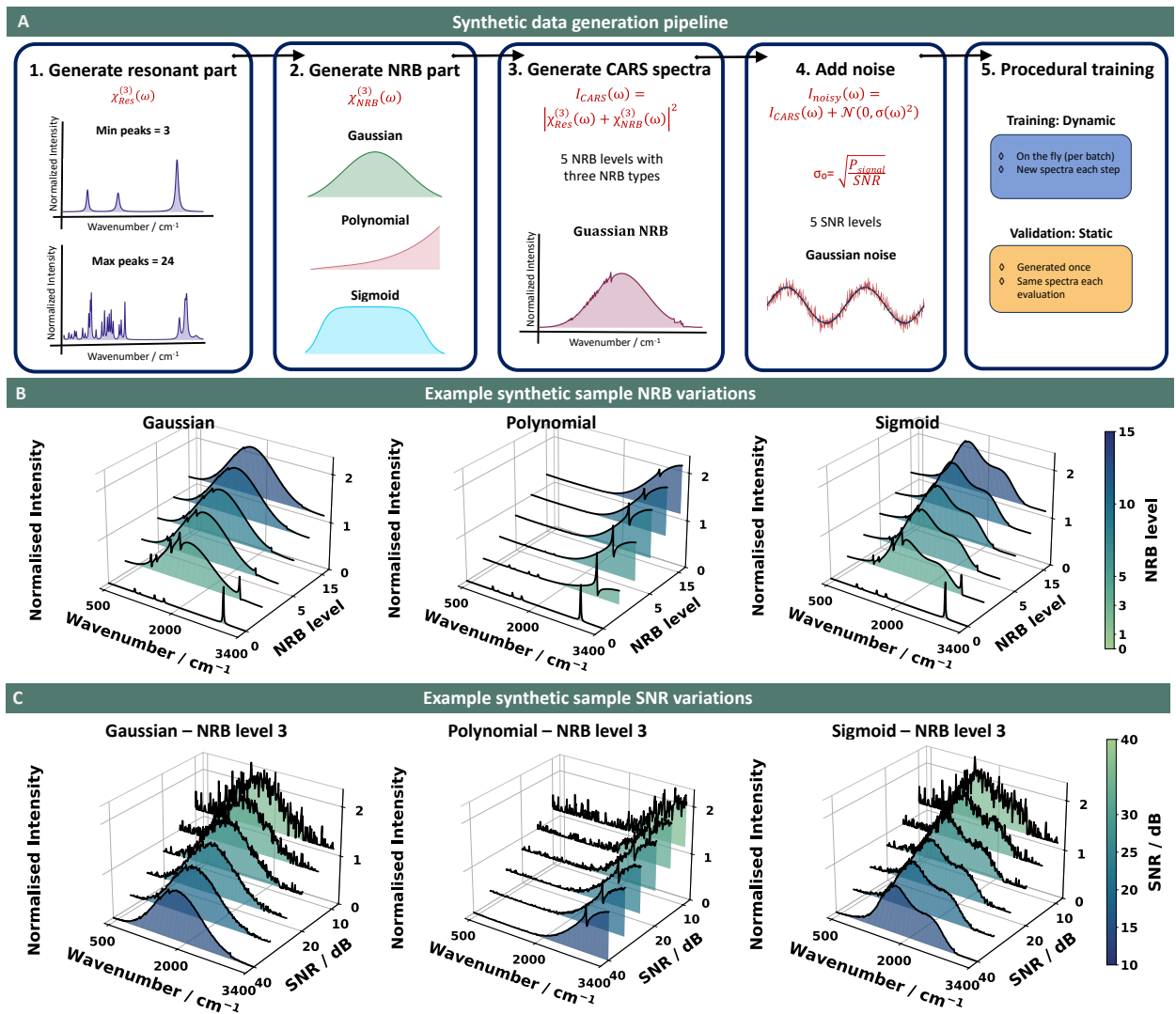}
\caption{Synthetic BCARS spectra used for training and validation. (A)~Each spectrum is generated from a resonant model based on reference band assignments and measurement standards. The solvent band set is kept fixed, while the background, the noise level, and the spectral coverage are varied to create realistic spectral diversity. A randomly sampled NRB is added to produce smooth baselines and interference-like distortions, followed by Gaussian noise to cover a wide range of signal quality. Validation spectra are generated once and then kept fixed (orange). (B)~Examples of the three NRB families used in this study (Gaussian, polynomial, and sigmoid) shown at increasing NRB levels. (C)~Examples of noise levels at a fixed NRB setting (level~3), showing the progression from clean to noisy spectra.}\label{fig:datapipeline}
\end{figure}

To train the inverse model across a wide range of measurement conditions, synthetic BCARS spectra were generated procedurally during training rather than drawn from a fixed precomputed dataset, a strategy also used by prior-data fitted networks, which are trained on data drawn repeatedly from a prior~\cite{muller2022}. This choice reflects the asymmetry of the inverse problem: experimental nuisance factors such as NRB shape, NRB strength, and noise can vary substantially, whereas the resonant spectrum is governed by a lower-dimensional band set. The full procedural pipeline is summarized in Fig.~\ref{fig:datapipeline}.

Each synthetic sample was generated in a fixed sequence. First, a solvent-specific band set was selected, defined by a set of nominal Raman bands. Band centers were taken from a per-solvent reference peak table, the relative intensities in that table were used as the amplitudes, and the damping constants were derived from its linewidths, with all values held fixed across samples.
This parameterization tied the band set to measured molecular signatures rather than to synthetic values, so that, for repeated samples of a given solvent, any variation between generated spectra originated in the measurement model rather than in the underlying band set.

The resonant part of the third-order susceptibility was then constructed as a sum of $K$ complex Lorentzian modes,
\begin{equation}
{\chi^{(3)}_{\mathrm{Res}}(\omega) = \sum_{k=1}^{K} \frac{A_k\,\omega_k^{2}}{\omega_k^{2} - \omega^{2} - i\,\Gamma_k\,\omega},}\label{eq:chi3}
\end{equation}
where $\omega$ is the Raman shift in wavenumber units (cm$^{-1}$), $i$ is the imaginary unit, and $K$ is the number of bands in the selected band set (between 5 and 24 across the twelve band sets used for training). Band $k$ is described by an amplitude $A_k$ (arbitrary units), a center $\omega_k$, and a damping constant $\Gamma_k$ that sets the linewidth. These are the parameters collected in $\theta$ (Section~\ref{subsec1_2}). For an isolated band, $\operatorname{Im}\chi^{(3)}_{\mathrm{Res}}$ equals $A_k\omega_k/\Gamma_k$ at $\omega = \omega_k$ and, for $\Gamma_k \ll \omega_k$, is close to a Lorentzian with a full width at half maximum of $\Gamma_k$, so simulated peak heights scale with $\omega_k/\Gamma_k$ rather than reproducing the tabulated intensities directly. This imaginary part is the resonant spectrum that iPINN aims to recover. Equation~\eqref{eq:chi3} was evaluated on a uniform grid of 3{,}000 points between 200 and 3{,}500~cm$^{-1}$, and the result was normalized to a maximum modulus of one. The noise-free BCARS intensity was then computed as
\begin{equation}
{I(\omega) = \left|\chi^{(3)}_{\mathrm{Res}}(\omega) + \chi^{(3)}_{\mathrm{NR}}(\omega)\right|^{2},}\label{eq:intensity}
\end{equation}
where $\chi^{(3)}_{\mathrm{NR}}(\omega)$ is the non-resonant susceptibility, taken as real, that gives rise to the NRB. Its shape and its magnitude relative to $\chi^{(3)}_{\mathrm{Res}}$ were sampled as described below. No separate excitation profile or instrument response was applied. The variety of sampled NRB shapes is intended to cover the background shapes of measured spectra.

The NRB was sampled from three functional families, as illustrated in Fig.~\ref{fig:datapipeline}B. A Gaussian NRB was included to represent smooth, low-curvature envelope distortions, which are consistent with the broad system-response and smoothly varying background structures commonly encountered in BCARS modelling~\cite{danehy2003}. A sigmoidal family was included to represent monotonic or asymmetric roll-off across the spectral window. This choice is also consistent with prior studies in which the NRB was parameterized using one or two sigmoid functions~\cite{junjuri2022}.

A polynomial NRB of order 4--5 was included because polynomial baseline models are widely used in Raman preprocessing, and fourth-order polynomial NRB models have also been shown to improve synthetic CARS training performance relative to sigmoid-only parameterizations~\cite{junjuri2022}. For each spectrum, one NRB family was selected, its family-specific parameters were drawn from broad uniform priors, and the resulting background was scaled relative to the resonant signal before noise was added. The ranges of these priors, the peak table, and all other generator settings are provided with the public code (see Data and Software Availability). Representative examples of these NRB shapes at increasing NRB strengths are shown in Fig.~\ref{fig:datapipeline}B.

After NRB synthesis, signal-dependent Gaussian noise was added to the simulated spectrum. The noise was modelled as heteroscedastic and spectrally correlated, with an optional slow baseline drift, as a simplified stand-in for the signal-dependent noise of real detectors~\cite{foi2008}. The nominal SNR was sampled uniformly from 2 to 60~dB to cover both low-light acquisitions and higher-quality averaged measurements. Representative examples of the resulting degradation are shown in Fig.~\ref{fig:datapipeline}C. Each noisy spectrum was then divided by its maximum value.

For a given solvent, the band set (centers, amplitudes, and damping constants) was held fixed. In contrast, noise factors were resampled for every generation. Specifically, each spectrum received a newly sampled NRB profile, an NRB-to-signal ratio drawn log-uniformly from 1 to 10, and noise with a nominal SNR of 2--60~dB. By keeping the band set fixed while randomizing measurement-specific distortions, the training pipeline encouraged the model to separate invariant molecular information from acquisition-dependent artifacts, as illustrated in Fig.~\ref{fig:datapipeline}A--C.

 Because spectra were generated procedurally rather than read from a static dataset, each epoch exposed the model to new combinations of NRB shape, NRB strength, noise level, and spectral coverage. In approximately 30\% of samples a randomly placed coverage crop removed the bands falling outside the retained window, so that the number of active peaks also varied between samples. The generator also supplies the grouping information needed for the multi-view consistency loss: each sample is assigned a solvent index when its Lorentzian parameters are selected, and predictions sharing that index are grouped within the batch, allowing multiple corrupted variations of the same underlying solvent to be regularized toward a common predicted band set. In practice, this means that the network repeatedly encounters the same underlying physics under different nuisance conditions, but not the same spectrum twice.

To monitor learning, a fixed validation set of about 9{,}700 spectra spanning diverse NRB profiles and noise levels was prepared once before training and held out throughout optimization. Keeping validation static while the training spectra change at every step makes the learning curves interpretable and allows changes in validation performance to be attributed to model improvement rather than to resampling variability.

\subsection{Experimental Datasets and Evaluation Data}\label{subsec2_1_2}

The iPINN model was evaluated on two experimental datasets. The first was a public BCARS benchmark comprising five experimental spectra acquired under different chemical and instrumental conditions. As shown in Fig.~\ref{fig:reconstructions}A--B, the benchmark includes toluene measured in two-color (2C) and three-color (3C) BCARS configurations~\cite{vernuccio2023}, together with spectra of dimethyl sulfoxide (DMSO) extracted at multiple SNR levels from a hyperspectral image of poly(methyl methacrylate) (PMMA) beads in DMSO~\cite{vernuccio2024,vernuccio2024supp}.

\begin{figure}[!htbp]
\centering
\includegraphics[width=\figwidth]{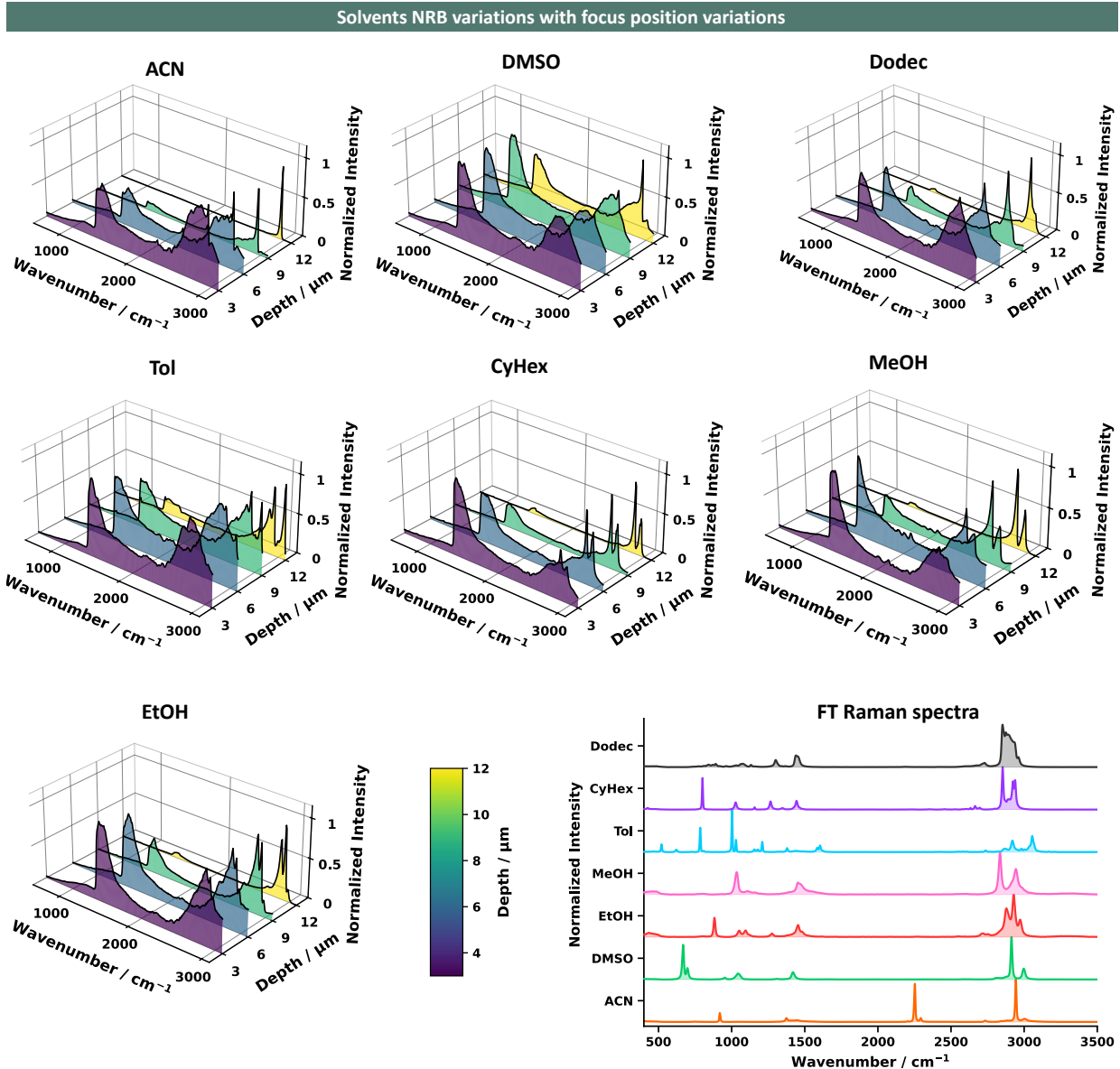}
\caption{Solvent benchmark dataset for experimental evaluation under NRB variability. (A)~Broadband BCARS spectra were acquired from seven solvents while sweeping the focal position (3--12~\textmu m) relative to the glass--solvent interface. This produced distinct NRB interference profiles and a progressive reduction in distortion with increasing depth. (B)~FT-Raman reference spectra were acquired at 1064~nm. BCARS measurements use a pump at 1032~nm and a Stokes spanning 1200--1600~nm (spectrally masked to separate 2C and 3C CARS regimes at 1350~cm$^{-1}$). This covers 600--3400~cm$^{-1}$ with a spectral resolution $<$10~cm$^{-1}$. Four focal positions per solvent yield 28 unique test spectra.}\label{fig:solventbench}
\end{figure}

To evaluate robustness under stronger and more variable NRB conditions, we also used a second experimental dataset acquired from seven reference solvents: acetonitrile (ACN), DMSO, ethanol (EtOH), methanol (MeOH), toluene (Tol), cyclohexane (CyHex), and n-dodecane (Dodec). BCARS spectra were collected on a broadband CARS system with a 1032~nm pump and a 1200--1600~nm Stokes range (spectrally masked to separate 2C and 3C CARS regimes), corresponding to a spectral window of 600--3400~cm$^{-1}$. This configuration produces distinct 2C and 3C CARS contributions separated at 1350~cm$^{-1}$, a transition clearly visible in spectra with elevated NRB levels. To introduce controlled variation in NRB strength, the focal position was swept from 3 to 12~\textmu m relative to the glass--solvent interface, producing four different NRB-to-resonant signal ratios for each solvent. Examples from this dataset are shown in Fig.~\ref{fig:solventbench}A.

This focus sweep creates a physically meaningful gradient of difficulty. At shallow focal positions, the interface produces a stronger and more complex non-resonant contribution, leading to heavily distorted BCARS spectra. As the focus is moved deeper into the liquid, the NRB contribution decreases, and the spectra become progressively cleaner. As shown in Fig.~\ref{fig:solventbench}A, this allows the model to be tested across a continuous range of background conditions rather than under a single fixed acquisition setting. Fourier-transform Raman (FT-Raman) reference spectra, shown in Fig.~\ref{fig:solventbench}B, were used to evaluate spectral fidelity.

Both datasets consist of solvent spectra, for which measured FT-Raman reference spectra allow a quantitative comparison over the shared spectral range. For heterogeneous biological samples, a comparable evaluation would require spatially matched reference measurements with comparable spatial and spectral resolution. Earlier experimental evaluations of DL-based NRB removal used, for example, three or four CARS spectra, each of a different sample~\cite{junjuri2022,junjuri2023}. The present evaluation comprises 33 experimental spectra in two datasets, 28 of them from seven solvents, with four NRB-to-resonant signal ratios per solvent.

\subsection{Model Architecture}\label{subsec2_1_3}

\begin{figure}[!htbp]
\centering
\includegraphics[width=0.83\figwidth,keepaspectratio]{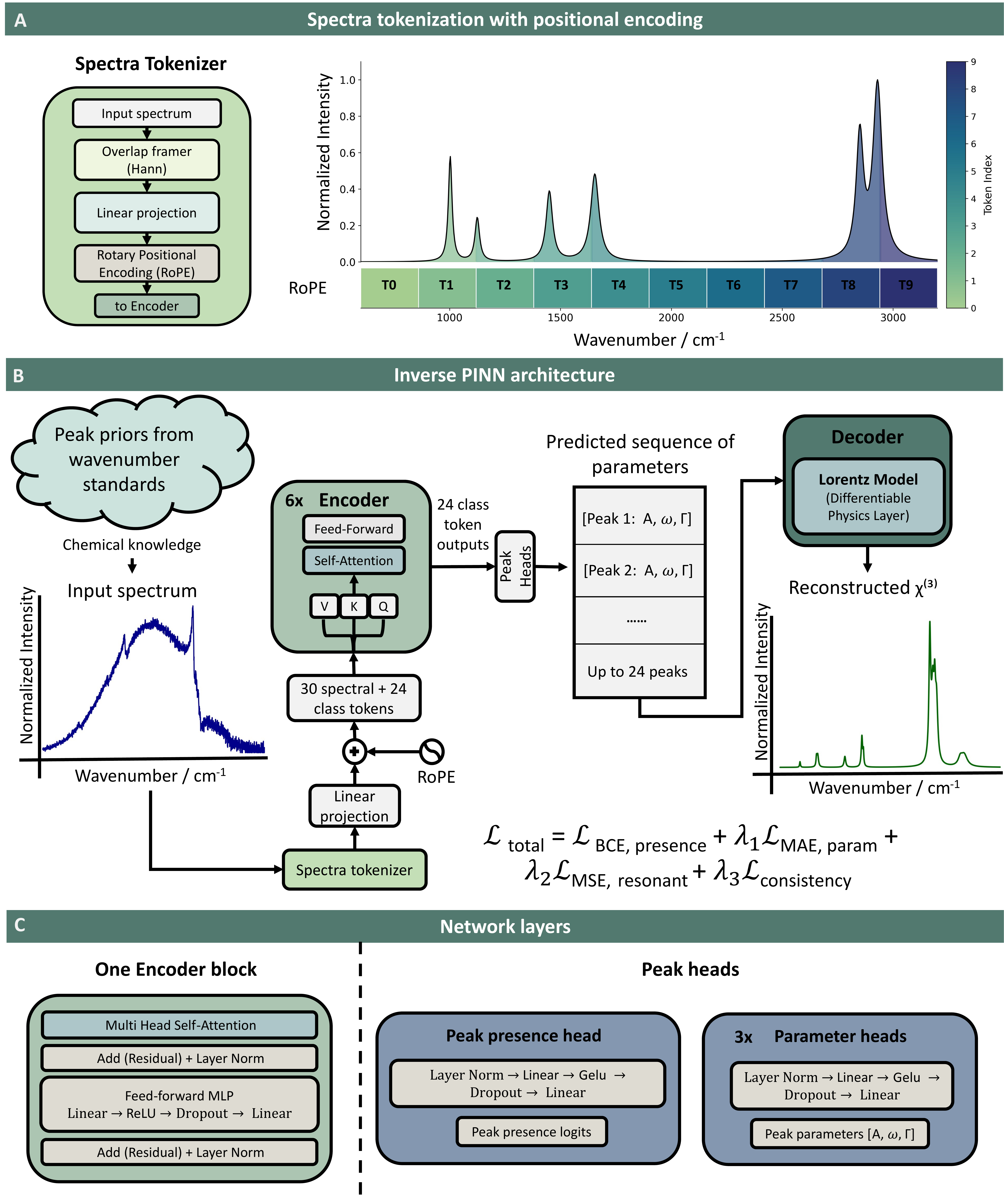}
\caption{Physics-informed architecture for inverse BCARS analysis. (A)~The input BCARS spectrum is divided into overlapping Hann-windowed segments, each treated as a spectral token. The tokens are linearly projected into the model's embedding space, and a rotary positional encoding then injects spectral order. The color gradient indicates token position along the wavenumber axis. (B)~A bank of learnable class tokens is concatenated to the spectral tokens, and the combined sequence is processed by a stack of Transformer encoder blocks. The class-token outputs are routed to peak heads, which emit a presence indicator and the corresponding Lorentzian parameters for each candidate peak. The predicted band set is passed to a differentiable physics layer based on the analytical Lorentz model. This layer is a parameter-free decoder that reconstructs the resonant spectrum from the predicted parameters. Wavenumber-standard peak priors inform synthetic training-data generation but are not used at inference. (C)~Internal structure of the encoder block and the prediction heads. Each encoder block combines multi-head self-attention with a feed-forward network, and both have residual connections and layer normalization. The presence head outputs one logit per slot, and the three parameter heads predict $A_k$, $\omega_k$, and $\Gamma_k$. The heads differ in depth to reflect the dynamic range of each parameter.}\label{fig:architecture}
\end{figure}

The proposed model (Fig.~\ref{fig:architecture}) predicts Lorentzian peak parameters from a BCARS spectrum and constrains those predictions with a differentiable physics layer, so that phase retrieval and parameter estimation are performed jointly. Because the measured CARS signal arises from coherent mixing between resonant and non-resonant contributions~\cite{cheng2004,junjuri2022}, the network is required to recover a physically meaningful band set, not only to reproduce the observed intensity. The model is an encoder--decoder system. The encoder, which maps the spectrum to a set of peak slots, is a transformer~\cite{vaswani2017}, whereas the decoder is the analytical Lorentz forward model itself, which is differentiable but parameter-free. It evaluates Eq.~\eqref{eq:chi3} for the predicted band set and returns its imaginary part, normalized by the maximum modulus, as the resonant spectrum. Because it is differentiable with respect to the predicted parameters, the gradient of the Lorentz-model loss (Section~\ref{subsec2_1_4}) reaches the encoder. The encoder is therefore trained end-to-end by a reconstruction signal in resonant-spectrum space, not by parameter regression alone.

A 3{,}000-point input spectrum is partitioned into 30 overlapping windows of length $L = 300$ with hop $H = 96$, yielding an overlap of approximately 68\% (Fig.~\ref{fig:architecture}A). Reflect-padding fills the right-most window, and each frame is multiplied by a Hann taper before embedding to suppress spectral leakage at token boundaries~\cite{lu2021}. Positions between approximately 410 and 3{,}380~cm$^{-1}$ fall within three or four windows, and positions closer to either end of the grid within one or two. Away from the ends, a band near the edge of one window therefore lies well inside a neighboring one.
Each Hann-weighted token is mapped into a 512-dimensional space by a single linear projection. The same windowed-token strategy has been used for transformer models in other spectral domains, including music audio analysis~\cite{lu2021}. A rotary positional encoding (RoPE) then injects spectral order into the token embeddings as a rotation rather than as an additive code~\cite{su2024}, which preserves relative position information across the depth of the encoder.

To allow several peaks to coexist within the same spectral region, a bank of 24 learnable class tokens, randomly initialised and deliberately without positional codes, is concatenated to the patch sequence (Fig.~\ref{fig:architecture}B). Withholding the positional code forces each class token to acquire its role through attention rather than through a fixed spectral position, so the same slot can bind to a peak at any wavenumber. Each class token then acts as a query for one candidate vibrational mode, while the 30 patch tokens supply the evidence. The resulting 54-token tensor is processed by six standard Transformer encoder blocks with post-norm layer normalization~\cite{vaswani2017}, so each class token can aggregate evidence from any region of the spectrum, and from the other class tokens, through joint self-attention. Each block has eight attention heads of head dimension 64, feed-forward width 1024, rectified linear unit (ReLU) activation, and dropout 0.1. After the final encoder layer, the patch outputs are discarded and only the 24 class-token outputs are read out. This design lets several peaks come from the same spectral region without competing for a slot: three closely spaced solvent modes can be predicted by three class tokens that all attend to the same patch. This behaviour is important in the crowded fingerprint range below 1{,}700~cm$^{-1}$. The number of slots, 24, equals the number of bands in the largest band set used for training (4-acetamidophenol). For every spectrum, the network returns all 24 slots, each with a presence logit and three parameters. During training, the bands of each band set are sorted by wavenumber and the $k$-th band is assigned to slot $k$. Slots without a band have a presence target of zero and are excluded from the parameter loss.

The 24 class-token outputs are routed in parallel to four task-specific multilayer perceptrons (Fig.~\ref{fig:architecture}C). For each candidate peak, the heads emit a presence logit and the three Lorentzian parameters $(A_k, \omega_k, \Gamma_k)$. The four heads are asymmetric by design. The amplitude head is the deepest, with two hidden layers of widths 1{,}024 and 512 and Gaussian error linear unit (GELU) activation, because peak amplitudes span the widest dynamic range and are the most sensitive to NRB-induced bias. The width head is the lightest, a single hidden layer of width 256 with ReLU activation, because the empirical $\Gamma$ distribution is narrow and an over-parameterized head tends to overshoot. The center and presence heads share a common two-layer design with GELU activation. All four heads emit raw, unconstrained outputs. The physical bounds $A_k \geq 0$ and $\Gamma_k > 0$ are not enforced inside the heads but downstream, when the predicted parameters enter the $\chi^{(3)}$ model. This separation simplifies the regression gradients and confines all hard physical constraints to a single place, the Lorentz-model loss described next.

\subsection{Training Objectives}\label{subsec2_1_4}

Each of the four prediction heads contributes a supervision signal. The presence head is supervised by a binary cross-entropy loss $\mathcal{L}_{\mathrm{BCE}}$ against a binary target that flags which slots correspond to real peaks. The three regression heads are supervised by a mask-weighted mean absolute error (MAE) loss $\mathcal{L}_{\mathrm{param}}$, with the mask ensuring that errors are computed only over slots flagged as real peaks. Within $\mathcal{L}_{\mathrm{param}}$ the per-parameter weights are 1.5 for amplitude, 2.0 for center, and 1.0 for width, which gives the center regression slightly more weight during training without overwhelming the amplitude and width signals. For a batch of $B$ spectra, with $\hat p_{bk}$ the predicted presence probability of slot $k$ in spectrum $b$ and $m_{bk}\in\{0,1\}$ its target, the two terms are
\begin{equation}
{\mathcal{L}_{\mathrm{BCE}} = -\frac{1}{24B}\sum_{b=1}^{B}\sum_{k=1}^{24}\left[m_{bk}\ln\hat p_{bk} + \left(1-m_{bk}\right)\ln\left(1-\hat p_{bk}\right)\right]}
\label{eq:lbce}
\end{equation}
and
\begin{equation}
{\mathcal{L}_{\mathrm{param}} = \frac{\sum_{b,k} m_{bk}\sum_{q} w_q\left|\hat\theta_{bkq}-\theta_{bkq}\right|}{\left(\sum_{b,k} m_{bk}\right)\sum_{q} w_q},}
\label{eq:lparam}
\end{equation}
where $\theta_{bkq}$ is the target value of parameter $q\in\{A,\omega,\Gamma\}$ scaled to $[0,1]$, $\hat\theta_{bkq}$ is the network output on the same scale, and $w_q$ are the per-parameter weights given above.

The third term is a physics-informed reconstruction loss, which we refer to as the Lorentz-model loss. At every training step, the predicted parameters are clipped to $[0,1]$, inverse-scaled to physical units and substituted into Eq.~\eqref{eq:chi3}, evaluated on the 3{,}000-point grid of Section~\ref{subsec2_1_1}. The reconstructed susceptibility is divided by its maximum modulus, as for the simulated target, and the $L_1$ distance between its imaginary part and the ground-truth $\operatorname{Im}\chi^{(3)}_{\mathrm{Res}}$ is back-propagated into the encoder. This Lorentz-model loss is
\begin{equation}
{\mathcal{L}_{\mathrm{Lorentz}} = \frac{1}{BN}\sum_{b=1}^{B}\sum_{n=1}^{N}\left|\frac{\operatorname{Im}\hat\chi_b(\omega_n)}{\max\!\left(\max_{n'}\left|\hat\chi_b(\omega_{n'})\right|,\,10^{-4}\right)}-y_b(\omega_n)\right|,}
\label{eq:llorentz}
\end{equation}
where $\hat\chi_b$ is Eq.~\eqref{eq:chi3} evaluated with the predicted parameters of the slots selected as described below, $y_b$ is the simulated resonant spectrum of spectrum $b$ normalized in the same way, $\omega_n$ is the $n$-th point of that grid, and $N = 3{,}000$. Because $\chi^{(3)}$ is rebuilt analytically from the predicted $(A_k, \omega_k, \Gamma_k)$ at every step, the encoder is supervised against the full spectral signature implied by its predictions, not only against the scalar parameter values. The slots used to assemble $\chi^{(3)}$ are those whose predicted presence probability exceeds 0.5. This selection is not differentiable, so the gradient of the Lorentz-model loss reaches only the parameter outputs of the selected slots, and the presence head is trained by the binary cross-entropy term alone.

The fourth term is a multi-view solvent-consistency loss $\mathcal{L}_{\mathrm{consistency}}$. For each underlying band set, several BCARS realizations are generated that share this band set but differ in NRB shape, NRB strength, and noise, drawn from the three NRB patterns described in Section~\ref{subsec2_1_1} (Fig.~\ref{fig:intro}B), and the model is required to make consistent predictions across all of them. This formulation follows the consistency-based family of semi-supervised objectives~\cite{sohn2020}, with the augmentation set restricted to realisations that are physically equivalent in terms of the underlying chemistry. The loss therefore penalizes only those changes in prediction that the underlying physics says should not occur. With $s(b)$ the solvent index of spectrum $b$ and $\bar\theta_{skq}$ the mean prediction for slot $k$ and parameter $q$ over the spectra in the batch with index $s$, the loss is
\begin{equation}
{\mathcal{L}_{\mathrm{consistency}} = \frac{\sum_{b,k,q} g_b\, m_{bk}\left|\hat\theta_{bkq}-\bar\theta_{s(b)kq}\right|}{3\sum_{b,k} g_b\, m_{bk}},}
\label{eq:lcons}
\end{equation}
where $g_b = 1$ if at least two spectra in the batch share the index $s(b)$ and $g_b = 0$ otherwise, and the factor 3 counts the three parameters per slot. It compares predictions only with each other, not with target parameter values.

The total objective combines the four terms with two warm-up schedules. It reads
\begin{equation}
{\mathcal{L} = \mathcal{L}_{\mathrm{BCE}} + \mathcal{L}_{\mathrm{param}} + \lambda_{\mathrm{L}}(e)\,\mathcal{L}_{\mathrm{Lorentz}} + \lambda_{\mathrm{c}}(e)\,\mathcal{L}_{\mathrm{consistency}},}
\label{eq:ltotal}
\end{equation}
where $e$ is the training epoch. The physics weight $\lambda_{\mathrm{L}}$ is introduced after a 15-epoch warm-up and ramped linearly to 0.25 over a further 40 epochs. The consistency weight $\lambda_{\mathrm{c}}$ is introduced after a 45-epoch warm-up and ramped to 0.005 over a further 90 epochs. The regression heads are therefore allowed to stabilize before the physics and consistency terms begin to shape the predicted band set. In earlier runs without this delay, gradients from a mis-specified $\chi^{(3)}$ pushed the heads into poor local minima. Optimisation uses AdamW, the variant of Adam with decoupled weight decay~\cite{loshchilov2017}. The base learning rate is $10^{-4}$, reduced by a factor of 100 for the class-token bank and the prediction heads to stabilize their joint adaptation with the deeper backbone, under a cyclic schedule of 10 epochs up and 15 epochs down. The batch size is 512, gradients are clipped at unit $L_2$ norm, and an exponential moving average of the parameters (decay 0.999) is maintained on a separate copy. These averaged weights are used at evaluation, while the live optimizer state continues to drive training. Training is performed in mixed precision to reduce memory footprint and accelerate the encoder pass, with one exception: the $\chi^{(3)}$ reconstruction path is forced to single precision because the squared wavenumbers in Eq.~\eqref{eq:chi3} (up to approximately $1.4 \times 10^{7}$~cm$^{-2}$ for centers at the 3{,}700~cm$^{-1}$ upper bound of the parameter scaling) exceed the half-precision range. A lower bound on the normalizing maximum keeps the normalization finite when no slot is selected. Each epoch consists of 3{,}000 steps of 512 newly generated spectra (about 1.5 million spectra), and no training spectrum is used twice. Training runs for 300 epochs, and the checkpoint with the lowest unweighted parameter MAE (Eq.~\eqref{eq:lparam} with all $w_q = 1$) on the validation set is kept. All training uses the procedurally generated spectra from Section~\ref{subsec2_1_1} mixed with a combination of NRB patterns at SNRs down to 2~dB, whereas the experimental BCARS data described in Section~\ref{subsec2_1_2} are reserved for zero-shot evaluation only.

\section{Results}\label{sec3}

\subsection{Generalization across the BCARS Phase Retrieval Benchmark}\label{subsec3_1_1}

We benchmarked iPINN against five published deep-learning models for direct CARS-to-Raman regression: SpecNet, a bidirectional long short-term memory network (BiLSTM), a convolutional network with gated recurrent units (CNN-GRU), a generative adversarial network (GAN), and VECTOR~\cite{valensise2020,junjuri2023,vernuccio2024,wang2022vector}. Each baseline was retrained from its authors' public code on spectra from the generator described in Section~\ref{subsec2_1_1}, with the same band sets, NRB families, and noise settings as iPINN, and with the training schedule reported in its original publication. Classical KK and maximum entropy retrieval were not included in this comparison, because they typically require a measured NRB reference and dataset-specific parameter choices (Section~\ref{subsec1}), whereas iPINN and the five baselines take only the measured BCARS spectrum as input. The benchmark comprises five experimental spectra drawn from two solvents under different acquisition conditions, and six models, including iPINN, were compared on these spectra. All model outputs were max-normalized before metric computation, so that comparisons are not biased by differences in absolute spectral scale.

\begin{figure}[!htbp]
\centering
\includegraphics[width=\figwidth]{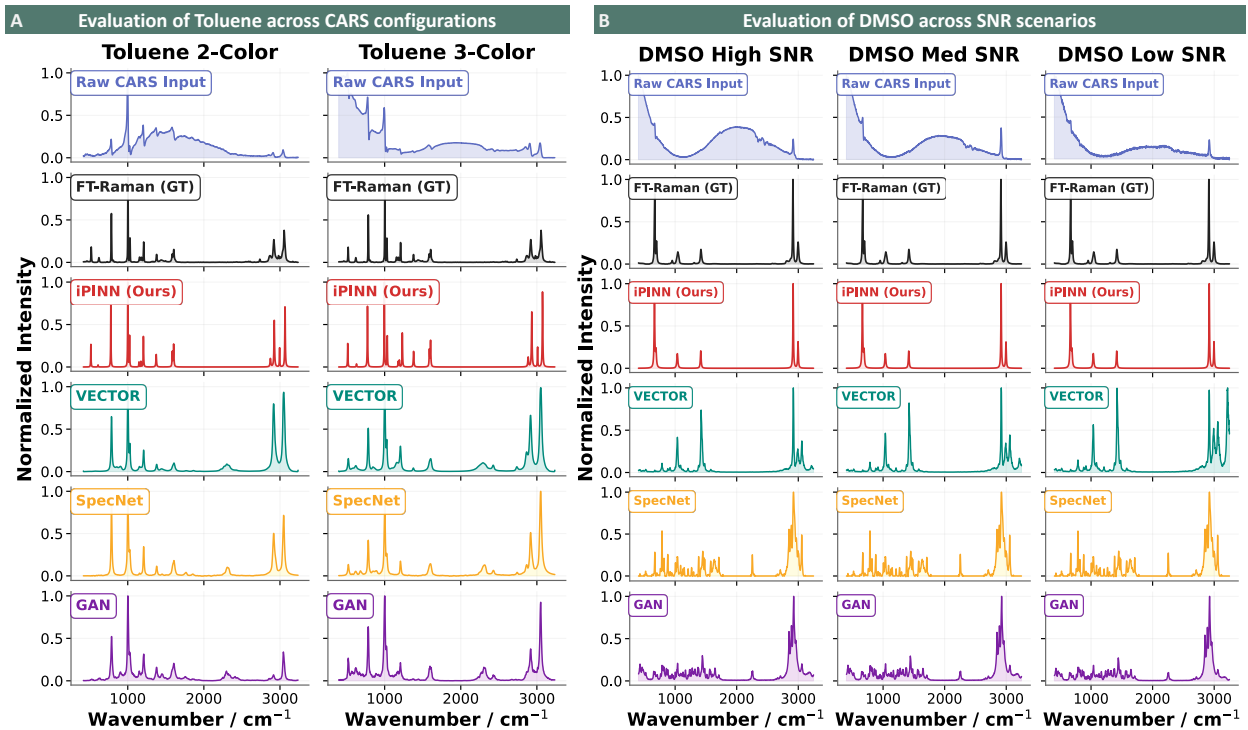}
\caption{Reconstruction performance on the public BCARS phase-retrieval benchmark. Each row of the spectral panels shows, from top to bottom, the raw CARS input, the FT-Raman reference (ground truth), and the spectra reconstructed by iPINN, VECTOR, SpecNet, and GAN, which are the three best-performing baselines from Figure~\ref{fig:quantitative}A. (A)~Toluene reconstructions for the 2C and 3C BCARS configurations. (B)~DMSO reconstructions at high, medium, and low SNR, using spectra extracted from the hyperspectral image of PMMA beads in DMSO.}\label{fig:reconstructions}
\end{figure}

Figure~\ref{fig:reconstructions}A shows the toluene reconstructions for the 2C and 3C BCARS configurations. In the raw CARS spectra, the broad NRB dominates the signal and the vibrational features appear as distorted asymmetric bands, whereas the FT-Raman reference shows the same features as sharp peaks on a flat baseline. iPINN recovered the major toluene peaks with the lowest absolute error in both configurations: MAE 0.019 for 2C and 0.025 for 3C.

Three properties of the iPINN reconstruction are visible in Fig.~\ref{fig:reconstructions}A. The model recovers the expected number of major peaks without introducing false positives. The silent region between 1700 and 2800~cm$^{-1}$ stays close to zero, with no spurious peaks, residual NRB curvature, or oscillatory artifacts. The reconstructed bands are smooth and symmetric, and the baseline between neighbouring peaks returns close to zero even in the congested low-wavenumber fingerprint region, particularly in the 500--800~cm$^{-1}$ range, where the ring-deformation modes near 521, 622, and 785~cm$^{-1}$ are closely spaced.

VECTOR, SpecNet, and GAN all recovered the dominant toluene peaks, but each retained visible artifacts. VECTOR preserved peak positions but showed an elevated background across the mid-wavenumber range. SpecNet produced broader line shapes and a non-zero baseline. GAN introduced low-amplitude oscillations between major peaks, especially in the 1200--1500~cm$^{-1}$ region. In all three baselines, the silent region remained contaminated by low-level residual structure. iPINN itself showed a tendency to overestimate the relative intensity of the C--H stretching cluster around 3000--3100~cm$^{-1}$ compared with the ring-breathing mode at 1003~cm$^{-1}$. VECTOR and SpecNet showed the same bias to a similar or larger degree.

Figure~\ref{fig:reconstructions}B evaluates DMSO under three SNR conditions, using spectra extracted from individual pixels of the hyperspectral image of PMMA beads in DMSO. The raw CARS inputs differ markedly from the toluene examples: the NRB dominates the full spectral envelope and the DMSO resonances appear only as weak inflections on a smooth background, which makes phase retrieval harder because the resonant contribution is small relative to the NRB. iPINN reconstructed the DMSO resonant spectrum consistently as SNR decreased, with MAE values of 0.011 (high SNR), 0.011 (medium SNR), and 0.012 (low SNR). Six expected vibrational bands were recovered: the symmetric C--S stretch at 670~cm$^{-1}$, the antisymmetric C--S stretch at 701~cm$^{-1}$, the S=O stretch near 1043~cm$^{-1}$, the CH$_{3}$ deformation at 1420~cm$^{-1}$, and the C--H stretching bands at 2912 and 2996~cm$^{-1}$. Across the three noise levels, peak positions and relative intensities stayed close to the FT-Raman reference and the silent region stayed near baseline. The baseline models showed more residual background and spurious low-level features, especially in weak-intensity regions.

\begin{figure}[!htbp]
\centering
\includegraphics[width=0.98\figwidth]{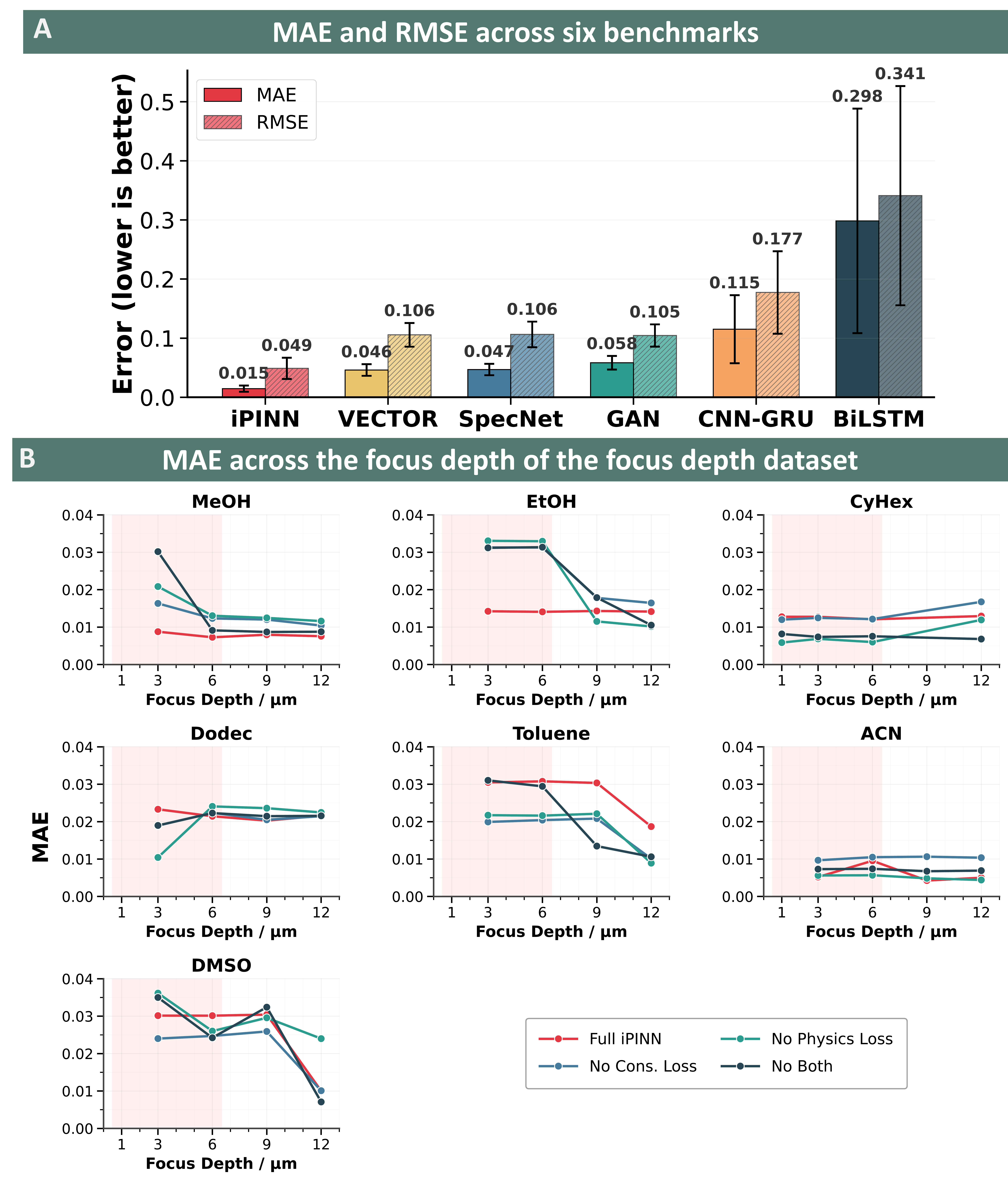}
\caption{Quantitative evaluation across the public benchmark and the focus-depth dataset. (A)~Mean MAE and RMSE for iPINN and five published baselines (VECTOR, SpecNet, GAN, CNN-GRU, BiLSTM) each averaged over the five benchmark spectra, with error bars of $\pm 1$ standard deviation across spectra. The in-panel heading ``six benchmarks'' refers to the six compared models. iPINN has the lowest mean error and the smallest variation. (B)~MAE as a function of focus depth (3, 6, 9, and 12~\textmu m) for the full iPINN and three ablation variants (no-Cons.\ loss, no-Physics loss, no-Both) on the focus-depth dataset, plotted separately for each of the seven solvents. The shaded band marks the strong-NRB depth range (1--6~\textmu m). Lower MAE indicates better reconstruction.}\label{fig:quantitative}
\end{figure}

Figure~\ref{fig:quantitative}A summarizes, for each of the six models, the MAE and the root-mean-square error (RMSE) averaged over the five benchmark spectra. iPINN achieved the lowest mean MAE (0.0156) and the lowest mean RMSE (0.0489), with the smallest error-bar spread across spectra. The next-best baseline was VECTOR (MAE 0.046), about three times higher than iPINN. SpecNet (0.047) and GAN (0.058) followed, while CNN-GRU (0.115) and BiLSTM (0.298) showed substantially larger errors and wider variation across spectra. To keep the spectral comparisons in Fig.~\ref{fig:reconstructions} readable, we plot iPINN alongside the three best-performing baselines from this ranking: VECTOR, SpecNet, and GAN.

\subsection{Stability across NRB Conditions on the Focus Depth Dataset}\label{subsec3_1_2}

The public benchmark above tests robustness to changes in solvent, excitation scheme, and SNR, but each spectrum is acquired under a single fixed measurement geometry. To examine robustness against systematic variation in NRB strength, we evaluated iPINN on the focus-depth dataset described in Section~\ref{subsec2_1_2}, where spectra were acquired at four focus depths (3, 6, 9, and 12~\textmu m) for seven solvents. Shallower focus produces a stronger NRB contribution from the glass--solvent interface. The shaded band in Fig.~\ref{fig:quantitative}B marks the strong-NRB depths. To quantify stability, we report the coefficient of variation (CV) of MAE across the four depths, defined as the standard deviation divided by the mean. A lower CV indicates more depth-invariant performance.

Figure~\ref{fig:quantitative}B plots the per-depth MAE for the full iPINN (red) on all seven solvents. For five of the seven solvents (MeOH, EtOH, CyHex, Dodec, ACN), the red line stays close to flat across the depth sweep. At 12~\textmu m, the model reached MAE values of 0.005 for ACN, 0.007 for DMSO, 0.008 for MeOH, and 0.013 for CyHex. Going from 12~\textmu m to 3~\textmu m, the error grew only slightly in the easier cases, from 0.005 to 0.006 for ACN and from 0.008 to 0.010 for MeOH. CyHex was the most depth-insensitive case in absolute terms (CV 2.6\%, Table~\ref{tab:ablation_cv_mae}), and EtOH had the lowest CV anywhere in the ablation table at 0.6\%, so reconstruction accuracy was effectively independent of focus depth for these two solvents.

The two cases with weaker depth invariance are toluene and DMSO. Tol MAE held near 0.030 at 3, 6, and 9~\textmu m and dropped to roughly 0.018 at 12~\textmu m, giving a CV of 18.6\%. DMSO showed a similar pattern, with a clear improvement only at 12~\textmu m and a CV of 34.7\%. In both cases the residual depth dependence is dominated by a single deeper-focus value rather than by a steady decline through the sweep.

\subsection{Ablation Study: Contributions of Physics and Consistency Losses}\label{subsec3_1_3}

To isolate the contribution of the two auxiliary losses, we trained three ablation variants alongside the full model: one without the consistency loss (no-Cons), one without the physics-informed Lorentz-model loss (no-Physics), and one without either loss (no-Both). Architecture, optimiser settings, and validation set were identical across the four variants, so any difference in behaviour reflects the loss configuration alone.

The four MAE-vs-depth curves in each panel of Fig.~\ref{fig:quantitative}B show that removing the physics loss is the dominant failure mode under strong NRB. For methanol, the no-Both variant climbs to MAE around 0.030 at 3~\textmu m before recovering at deeper focus, while the full iPINN stays near MAE 0.008 across the same range. For ethanol, all three ablation variants reach MAE around 0.032 at 3~\textmu m and remain elevated through 6~\textmu m, while the full iPINN holds near MAE 0.014 at every depth. The same general ordering, with the full iPINN closest to flat and the no-Both variant the most depth-sensitive, is repeated in the MeOH, EtOH, CyHex, and Dodec panels.

\begin{figure}[!htbp]
\centering
\includegraphics[width=\figwidth]{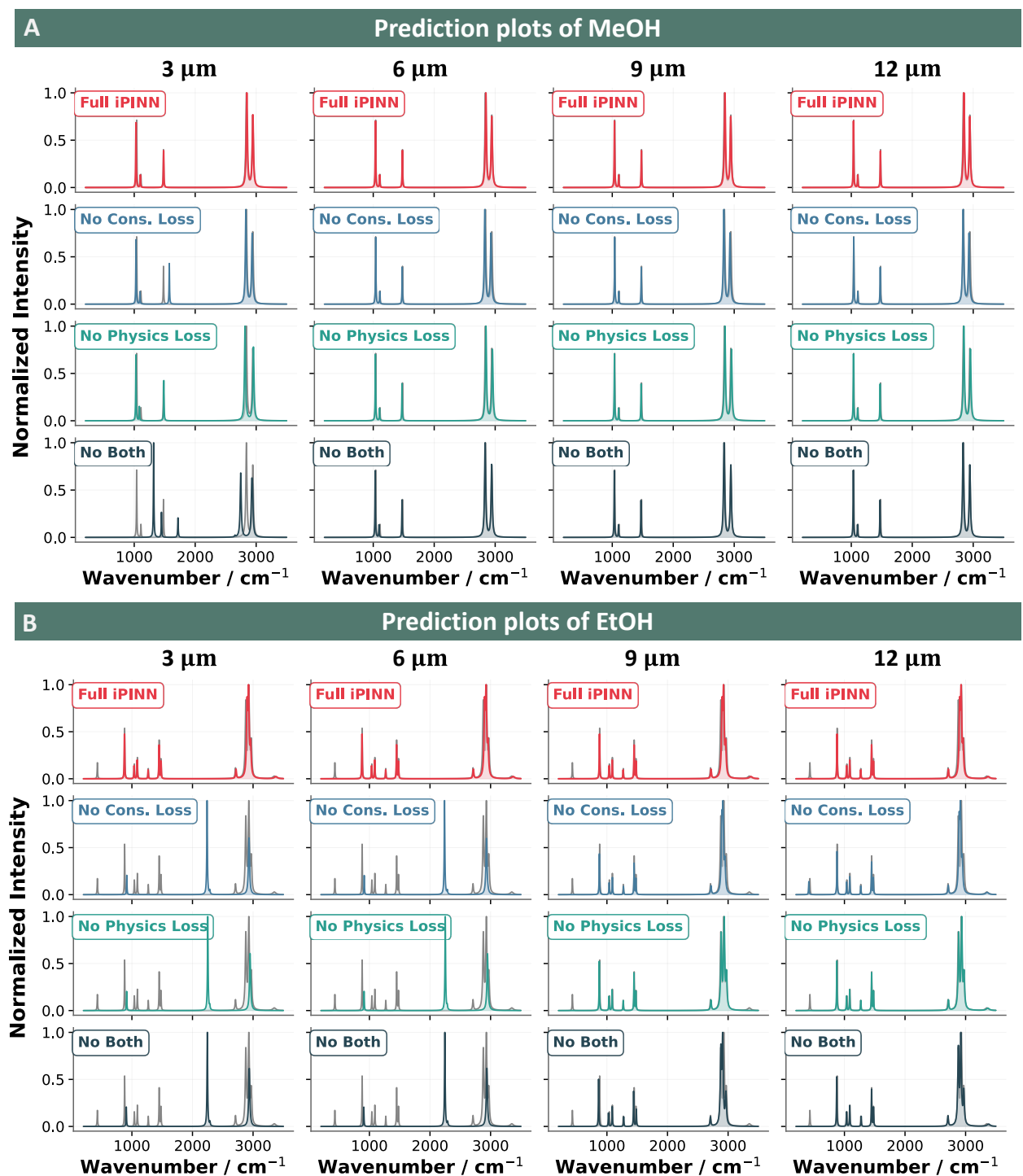}
\caption{Spectral reconstructions across focus depth for the full iPINN and the three ablation variants. Predicted resonant spectra at four focus depths (3, 6, 9, and 12~\textmu m, one per column) for the four model variants (rows), shown against the FT-Raman reference (grey). (A)~MeOH. (B)~EtOH. The full iPINN preserves peak shape and silent-region baseline across depths in both solvents, while the ablation variants deviate from the reference most strongly at 3 and 6~\textmu m.}\label{fig:ablationdepth}
\end{figure}

The corresponding spectral reconstructions for MeOH and EtOH are shown in Fig.~\ref{fig:ablationdepth}, at all four focus depths and for all four model variants, with the FT-Raman reference plotted in grey behind each prediction.

For MeOH (Fig.~\ref{fig:ablationdepth}A), the full iPINN row produced visually similar reconstructions across all four depths. The no-Both row degraded most strongly at 3~\textmu m, with broadened peaks, baseline distortion, and visible intensity in the silent region between the fingerprint and C--H stretching ranges. At 3~\textmu m, the average predicted intensity in non-peak regions was 5.1 times the ground-truth baseline for no-Both, compared with 1.2 times for the full iPINN. Across the depth sweep, the full-model MAE ranged from 0.0073 to 0.0088 (CV 7.1\%), while no-Both ranged from 0.0087 to 0.0302 (CV 65.0\%).

For EtOH (Fig.~\ref{fig:ablationdepth}B), the full iPINN row closely matches the FT-Raman reference at every depth, and the silent-region baseline remains clean. The three ablation rows visibly deviate from the reference at 3 and 6~\textmu m, with the deviation easing at 9 and 12~\textmu m. The CV values in Table~\ref{tab:ablation_cv_mae} quantify the gap: the full iPINN reaches CV 0.6\% on ethanol, the lowest value in the table, while no-Cons rises to 29.3\%, no-Both to 39.2\%, and no-Physics to 50.6\%. For EtOH, the physics loss is therefore the largest single change to remove, and the consistency loss provides additional stability beyond it.

A direct comparison between no-Cons and no-Both isolates the role of physics loss. At 3~\textmu m on methanol, no-Cons reached MAE 0.0163 while no-Both reached MAE 0.0302. In this case, the physics loss alone roughly halves the error under the strongest NRB. The consistency loss reduces variation in relative peak amplitudes across measurement conditions.

Table~\ref{tab:ablation_cv_mae} summarizes depth stability across all seven solvents. The full iPINN has the lowest CV in five of the seven solvents, with the strongest gains on MeOH (7.1\% versus 65.0\% for no-Both), EtOH (0.6\% versus 29.3\% for no-Cons), and CyHex (2.6\% versus 32.5\% for no-Physics). For ACN and DMSO the lowest CV is reached by an ablation variant rather than by the full model. In these two cases the absolute MAE values for the full iPINN are already small (around 0.005 for ACN and as low as 0.007 at 12~\textmu m for DMSO), and the CV is dominated by the gap between the deepest-focus value and the rest of the sweep, so a flatter but uniformly less accurate ablation variant can score a lower CV without producing better reconstructions. The two losses therefore behave as complementary regularizers under varying NRB strength, with the lowest CV across the dataset reached only when both are active.

\begin{table}[!htbp]
\caption{Ablation of iPINN loss components and comparison with baseline methods. The coefficient of variation (CV) of MAE is reported in percent (lower values indicate better stability).}
\label{tab:ablation_cv_mae}
\centering
\scriptsize
\setlength{\tabcolsep}{4pt}
\begin{tabular*}{\textwidth}{@{\extracolsep\fill}lccrrrrrrr}
\toprule
\multirow{2}{*}{Variant} & \multicolumn{2}{c}{Loss components} & \multicolumn{7}{c}{CV of MAE (\%) $\downarrow$} \\
\cmidrule(lr){2-3}\cmidrule(lr){4-10}
 & $\chi^{(3)}$ & Cons. & MeOH & EtOH & CyHex & Dodec. & Tol & ACN & DMSO \\
\midrule
\multicolumn{10}{@{}l}{\textbf{iPINN variants}} \\
Full iPINN & \checkmark & \checkmark & \textbf{7.1} & \textbf{0.6} & \textbf{2.6} & \textbf{5.0} & \textbf{18.6} & 34.5 & 34.7 \\
No Cons. loss & \checkmark & $\times$ & 17.1 & 29.3 & 14.8 & 6.1 & 25.1 & \textbf{3.6} & 30.4 \\
No Physics Loss & $\times$ & \checkmark & 25.6 & 50.6 & 32.5 & 28.0 & 30.1 & 10.3 & \textbf{16.0} \\
Both losses & $\times$ & $\times$ & 65.0 & 39.2 & 6.5 & 5.9 & 43.4 & 3.8 & 44.2 \\
\midrule
\multicolumn{10}{@{}l}{\textbf{Baselines}} \\
GAN & $\times$ & $\times$ & 22.8 & 46.9 & 42.6 & 11.9 & 57.9 & 18.1 & 43.2 \\
SpecNet & $\times$ & $\times$ & 29.5 & 44.1 & 49.9 & 7.2 & 37.2 & 54.5 & 36.7 \\
\bottomrule
\end{tabular*}
\end{table}

\section{Discussion and Future Outlook}\label{sec4}

The advantage of iPINN over the baseline models is structural: the network does not predict the spectrum directly. It infers peak positions, amplitudes, and widths, and the resonant spectrum is reconstructed from those parameters through the Lorentzian model. The reconstruction is therefore always a sum of smooth Lorentzian peaks, and regions without predicted vibrational activity return close to baseline. False peaks and residual background structure are suppressed by construction. The baseline models predict every spectral point independently, which leaves room for background artifacts and features unconnected to any molecular vibration. iPINN reached the lowest MAE on all five benchmark spectra. It also produced cleaner spectra in the silent region, and the MAE differed by only 0.001 between the high-SNR and low-SNR DMSO cases.

    The same parameter-prediction strategy changes how the NRB is handled. Classical KK methods face a circular dependency: the KK transform requires normalization by the NRB, which is itself unknown~\cite{liu2009}. Recent dual-decoder physics-informed networks address this by jointly estimating the NRB and the resonant spectrum in parallel branches, but the NRB still has to be reconstructed explicitly~\cite{vemuri2025}. Predicting $\theta$ directly removes this requirement. The physics loss compares the resonant spectrum reconstructed from $\theta$ with the simulated resonant spectrum, so no explicit NRB estimate is needed during training, and the NRB enters only through data generation.

The ablation study over four focus depths, seven solvents, and 28 spectra shows the contribution of each loss term. The physics loss suppresses false peaks under strong NRB. This is most visible for methanol at 3~\textmu m in Fig.~\ref{fig:ablationdepth}A, where removing the physics loss introduces spurious peaks and increases the MAE by 46\% relative to the full iPINN. The consistency loss controls the relative amplitudes of closely spaced peaks. For ethanol at 9~\textmu m in Fig.~\ref{fig:ablationdepth}B, the no-Physics variant gives a 36\% lower MAE than no-Both, indicating that recovering correct relative amplitudes in the C--H stretching cluster requires the consistency objective specifically. With both losses active, performance remains the most stable as the NRB changes: the full iPINN reaches the lowest CV in five of the seven solvents in Table~\ref{tab:ablation_cv_mae}, and the depth-MAE curves in Fig.~\ref{fig:quantitative}B stay flattest under this configuration.

Table~\ref{tab:ablation_cv_mae} also shows that the full model is not the best for every solvent. For ACN and DMSO, an ablation variant reaches a lower CV. The likely cause is the forward model itself: the Lorentzian sum captures peak positions accurately but does not yet reproduce the exact NRB envelope of every solvent under changing focus conditions. For solvents dominated by a small number of strong bands, or where the measured envelope departs from the idealized Lorentzian shape, small mismatches in modelled peak height or residual NRB curvature can shift with focus depth and reduce consistency, even when peak positions are recovered correctly. The full model is therefore best on average but not uniformly across all seven solvents.

This analysis points to two limitations of the present method. The first is the lower consistency seen for some solvents under changing background conditions, which reflects cases where the forward model cannot fully reproduce the measured envelope. A loss that penalizes relative amplitude errors between neighbouring peaks could address part of this, and a forward-modelling refinement step that adjusts NRB shape and peak heights while keeping positions fixed could close the rest, without disturbing the peak positions iPINN already predicts well. The second limitation is the fixed maximum of 24 peaks per spectrum. This is sufficient for common solvents such as toluene and DMSO, but biological samples with dense, overlapping vibrational bands may require more. Allowing the network to infer the number of active peaks from the input could broaden the method to more demanding chemical and biological imaging tasks.

The depth-invariant behaviour reported here is relevant in practice. Hyperspectral BCARS images contain pixel-to-pixel variation in effective NRB, and a model that stays stable under this variation could make spectra directly comparable across an image without per-pixel correction.

The twelve band sets used for training form the training library of iPINN, and all solvents evaluated here belong to it. The resonant spectrum of a sample is set by its molecular vibrations and does not change with the NRB of a measurement or with the detection noise. A phase-retrieval method should therefore return the same resonant spectrum, and for iPINN the same band set, for the same sample under different NRB and noise conditions. The focus-depth dataset tests this property at four NRB-to-resonant signal ratios for each solvent (Section~\ref{subsec2_1_2}), because the FT-Raman reference of a solvent is the same at every depth. The coefficients of variation in Table~\ref{tab:ablation_cv_mae} summarize the relative variation of the MAE across the four depths. In the benchmark, the same toluene FT-Raman reference is used for the spectra from two excitation schemes, and the same DMSO FT-Raman reference for the spectra at three SNR levels. The training data follow the same principle (Section~\ref{subsec2_1_1}).

BCARS images of tissues have been formed from marker bands, for example those of nucleic acids at 785~cm$^{-1}$, proteins at 1{,}004~cm$^{-1}$, and lipids at 2{,}850~cm$^{-1}$~\cite{camp2014}. These marker bands motivate extending iPINN to the band sets of selected biological constituents. This would require an expanded training library, retraining with enough slots for the combined band sets, and evaluation on mixtures and biological samples. As a first approximation, mixtures could be simulated by summing the susceptibilities of their components with sampled weights, assuming that the response of each component does not change on mixing. Several corrupted views of each sampled mixture would then form one group for the consistency loss. Inhomogeneous broadening with an approximately Gaussian profile could be represented by a complex Voigt response. This would add one Gaussian width per band and would require matching changes to the training generator and the predicted parameters.

The inverse-first formulation also suggests a structural change at the input side of the network. The current spectrum tokenizer uses a linear projection of overlapping Hann-windowed segments followed by rotary positional encoding (Fig.~\ref{fig:architecture}A). The projection learns its filters from data and has to discover during training that the relevant local features are Lorentzian. This prior could instead be encoded directly at the input by replacing the linear projection with a layer of parametric Lorentzian neurons, each of which outputs a Lorentzian response with a learnable amplitude, center, and width. This is the input-side counterpart of the differentiable Lorentz model at the output, where in both cases a learned approximation is replaced by a parametric form whose parameters carry physical meaning. With this prior built into the tokenizer, the network would rely less on training data to express the spectroscopic structure of the input, which could reduce the synthetic-to-real gap. Together with adaptive peak budgets and forward-model refinement, the method could then be extended to chemically heterogeneous and spectrally dense samples, including those encountered in label-free pathology and live-tissue imaging.

\section*{Acknowledgements}

This work is supported by the BMBF funding program Photonics Research Germany (13N15464 (LPI-BT1-IPHT), 13N15466 (LPI-BT1-FSU), and 13N15706 (LPI-BT2-FSU)) and is integrated into the Leibniz Center for Photonics in Infection Research (LPI). The LPI, initiated by Leibniz-IPHT, Leibniz-HKI, Friedrich Schiller University Jena, and Jena University Hospital, is part of the BMBF national road map for research infrastructures. This project has received funding from the European Union's Horizon 2020 research and innovation programme under grant agreement No.~101016923 (CRIMSON), and from the European Union's Horizon Europe research and innovation programme under grant agreement No.~101135175 (uCAIR). The authors gratefully acknowledge support from the Carl Zeiss Foundation through the project ``Sensorized Surgery'' (P2022-06-004), and from Deutsche Krebshilfe under Projekt-Nr.~70116281 (Arbor). This work was co-funded by the Deutsche Forschungsgemeinschaft (DFG, German Research Foundation)~--~441958208 (NFDI4Chem). We acknowledge support by the German Research Foundation Projekt-Nr.~512648189 and the Open Access Publication Fund of the Th\"uringer Universit\"ats- und Landesbibliothek Jena.

\section*{Author Contributions}

Conceptualization, R.T.V. and T.B.; methodology, R.T.V. and T.B.; software, R.T.V.; formal analysis, R.T.V.; investigation, R.T.V.; data curation, C.M., M.V., T.M.-Z.; writing, original draft preparation, R.T.V.; writing, review and editing, R.T.V., T.B., C.M., M.V., R.J., T.M.-Z., and J.P.; visualization, R.T.V.; supervision, T.B.\ and J.P.; project administration, T.B.; funding acquisition, T.B.\ and J.P. All authors have read and agreed to the published version of the manuscript.

\section*{Notes}

The authors declare no competing financial interest.

\section*{Data and Software Availability}

The public BCARS phase-retrieval benchmark used in this work is available from the cited sources. The focus-depth solvent dataset and the iPINN implementation are available at \url{https://git.photonicdata.science/ravi_vulchi/inverse_pinn_paper.git}.

\printbibliography

\end{document}